\documentclass[conference]{IEEETran}
\IEEEoverridecommandlockouts
\usepackage{cite}
\usepackage{amsmath,amssymb,amsfonts}
\usepackage{algorithmic}
\usepackage[ruled,vlined]{algorithm2e}
\usepackage{graphicx}
\usepackage{tabularx}
\usepackage{booktabs}
\usepackage{textcomp}
\usepackage{xcolor}
\def\BibTeX{{\rm B\kern-.05em{\sc i\kern-.025em b}\kern-.08em
    T\kern-.1667em\lower.7ex\hbox{E}\kern-.125emX}}
    
\usepackage{etoolbox}
\makeatletter
\patchcmd{\@makecaption}
  {\scshape}
  {}
  {}
  {}
\patchcmd{\@makecaption}
  {\\}
  {.\ }
  {}
  {}
\patchcmd{\@makecaption}
  {\centering}
  {}
  {}
  {}
\makeatother
\def\tablename{Table}

\graphicspath{{./figures/}}

\begin{document}

\title{Interpretable Diversity Analysis: Visualizing Feature Representations In Low-Cost Ensembles}

\author{\IEEEauthorblockN{Tim Whitaker}
\IEEEauthorblockA{\textit{Department of Computer Science} \\
\textit{Colorado State University}\\
Fort Collins, CO, USA \\
timothy.whitaker@colostate.edu \\[-3.0ex]}
\and
\IEEEauthorblockN{Darrell Whitley}
\IEEEauthorblockA{\textit{Department of Computer Science} \\
\textit{Colorado State University}\\
Fort Collins, CO, USA \\
whitley@cs.colostate.edu \\[-3.0ex]}
}
\maketitle

\begin{abstract}
Diversity is an important consideration in the construction of robust neural network ensembles. A collection of well trained models will generalize better if they are diverse in the patterns they respond to and the predictions they make. Diversity is especially important for low-cost ensemble methods because members often share network structure in order to avoid training several independent models from scratch. Diversity is traditionally analyzed by measuring differences between the outputs of models. However, this gives little insight into how knowledge representations differ between ensemble members. 
This paper introduces several interpretability methods that can be used to qualitatively analyze diversity.
We demonstrate these techniques by comparing the diversity of feature representations between child networks using two low-cost ensemble algorithms, Snapshot Ensembles and Prune and Tune Ensembles.
We use the same pre-trained parent network as a starting point for both methods which allows us to explore how feature representations evolve over time.
This approach to diversity analysis can lead to valuable insights and new perspectives for how we measure and promote diversity in ensemble methods.
\end{abstract}
 

\section{Introduction}

Ensemble learning has long been known to be an effective technique for improving generalization on machine learning tasks.
Instead of making predictions with a single model, ensembles train multiple independent models and combine their predictions together.
The combination of several models can help to reduce bias that a single model might have.
This idea of diversity is an important topic in ensemble learning as it is critical in building ensembles that generalize well \cite{sollich1995learning, kuncheva2003measures}.

As neural networks and datasets 
become larger, efficient algorithms have been introduced to reduce the computational cost of training several models from scratch.
In general, this is done by sharing information between ensemble members in some way.
For example, Snapshot Ensembles train a single model and save checkpoints of 
the model throughout time \cite{huang2017snapshot}.
Prune and Tune Ensembles train a single parent model and then spawn independent child networks by pruning random parameters from the parent \cite{whitaker2022prune} followed by fast retraining.
Diversity becomes especially important for these low-cost ensemble methods as information sharing invariably leads to higher correlation between predictions.

Many metrics have been introduced to measure diversity in classifier ensembles \cite{kuncheva2003measures}.
Most of these metrics involve analyzing the differences between output predictions of different members. 
While this approach enables easy comparisons between classifiers, output differences can vary greatly depending on the accuracy of individual members. For example, a group of random models can be highly diverse but have terrible performance and a group of perfectly accurate models can have terrible diversity but perform well. Because of this, output diversity metrics can be misleading when comparing the results of different ensemble methods.

Neural network interpretability continues to be 
important in machine learning research \cite{zhang2021interpretability}. This field is shifting the preconceived notions that deep neural networks act as black box models. By utilizing feature visualization and saliency attribution techniques, it becomes possible to explore \textit{how}, \textit{what}, and \textit{why} neural networks make decisions by visualizing the types of patterns that specific neurons within the network respond to.

These methods are especially 
critical for evaluating low-cost ensemble methods that share network structure, because we can meaningfully compare feature representations of the same neuron in two different networks and visualize how they evolve over time.
We explore this idea with Snapshot Ensembles and Prune and Tune Ensembles by using the same pre-trained Inception-V3 network as a starting point for both methods \cite{huang2017snapshot, whitaker2022prune}.
We generate child networks for each ensemble by training on ImageNet and we then create neuron visualizations for each neuron in each of the child networks.
We then quantify the representational diversity by measuring the similarity between the visualizations using well established perceptual hashing algorithms \cite{buchner2021imagehash}.

Our results indicate that Prune and Tune Ensembles display significantly more representational diversity than Snapshot Ensembles for each of the five benchmark perceptual hashing algorithms we used to measure similarity.
The unique network topology of Prune and Tune children encourages each ensemble member to learn unique patterns. 
This insight could be highly effective for building more robust ensemble techniques in the future.
Our experiments offer a new perspective for analyzing diversity.
The application of interpretability techniques can be invaluable for better understanding how to measure and promote diversity in low-cost ensembles.


\pagebreak

\section{Background}

\subsection{Low-Cost Ensemble Learning}

Several methods have been introduced to reduce the costs of training ensembles, which generally share either network structure, gradient information, or learned parameters among ensemble members.
This can be effective at reducing computational cost, as each network can leverage the knowledge learned from other members without needing to be trained from scratch.
However this invariably reduces diversity between members and can have significant impact on generalization performance \cite{kuncheva2003measures}.
In this paper, we explore diversity using two recent low-cost ensemble learning algorithms, Prune and Tune Ensembles and Snapshot Ensembles.



\subsection{Prune and Tune Ensembles}

Prune and Tune Ensembling (PAT) is an efficient low-cost ensemble learning method that leverages ideas from evolutionary and temporal ensembles in order to efficiently create accurate and diverse networks through pruning.

Prune and Tune Ensembles work by first training a single parent network, then spawning child networks by cloning and dramatically pruning the parent using random or anti-random sampling strategies, and finally fine tuning each of the child networks with a cyclic learning rate schedule for a small number of epochs.
Pruning and tuning a previously trained network has an extremely low computational cost. Because the parent is already optimized, the child networks converge with only a few epochs of additional training. One can easily create a parent and dynamically generate many diverse child networks at a cost that is only slightly more than training a single network.

As the child networks are all derived from an identical parent network and tuned for a small number of epochs, one would assume that the children would be highly correlated. However, results have shown that Prune and Tune Ensembles produce extremely accurate ensembles with good output diversity.
We aim to better explore the feature representations of these child networks in order to understand how sparsity affects diversity.



Anti-random pruning was introduced in Prune and Tune Ensembles as a technique for encouraging more diversity among pairs of child networks \cite{malaiya1995antirandom, whitaker2022prune}. Anti-random pruning creates \textit{mirrored} pairs of child networks, such that whenever we randomly prune the parent to create a child, a sibling is created that inherits the opposite set of parameters.

Consider a binary bit string $M = \{x_0, ..., x_n: x \in \{0, 1\}\}$, that is randomly generated with 50\% sparsity where $1$ represents parameters that we keep and $0$ represents parameters that are pruned. The anti-random network then is created by reversing the polarity of all the bits in the mask $M$, such that:
\[\hat{\theta}_1 = \theta \circ M ~~~  \mbox{and} ~~~ \hat{\theta}_2 = \theta \circ (1 - M)\]
where $\hat{\theta}_i$ are the parameters of child network $i$, $\theta$ are the parameters of the parent network and $\circ$ denotes the Hadamard product.

The resulting two child networks maximize the Cartesian Distance, $CD$, between the two binary bit masks $a$ and $b$, where $a = M$ and $b = 1 - M$.
\[
CD(a,b) = \sqrt{|a_1 - b_1| + ... + |a_n - b_n|}
\]
This process is repeated $N$ times to create an ensemble of size $2N$, where each child network has exactly 50\% sparsity.

\begin{figure}[t]
    \centering
    \includegraphics[width=\columnwidth]{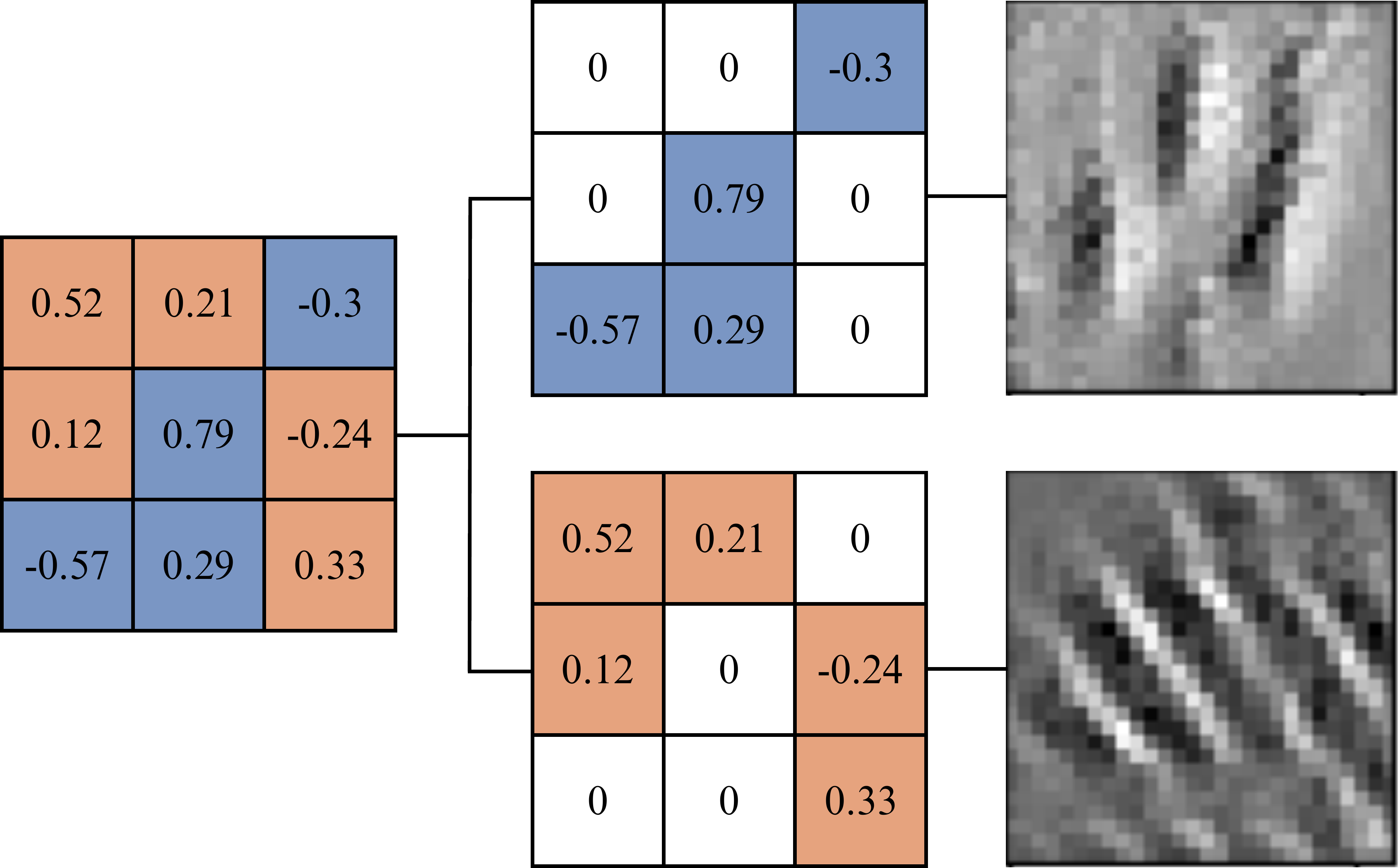}
    \caption{A parent filter split into two anti-random child filters. The resulting networks learn diverse feature representations as a result of their unique topologies.}
    \label{fig:my_label}
\end{figure}

In order to recover accuracy after pruning, each child undergoes a small tuning phase using a cyclic learning rate schedule. One-cycle tuning is a two phase schedule that inversely cycles between growth and decay for both learning rate and momentum \cite{smith2018superconvergence}. The first phase grows the learning rate from $\eta_{min}$ to $\eta_{max}$, while momentum decays from $\mu_{max}$ to $\mu_{min}$. The cycle then flips to decay the learning rate from $\eta_{max}$ to $\eta_{min}$ and the momentum from $\mu_{min}$ to $\mu_{max}$. Both learning rate and momentum are updated using cosine annealing.
\[
F(t) = \alpha_{0} + \frac{1}{2}(\alpha_{1} - \alpha_{0})
(1 + cos(\frac{t}{t_{max}} \pi))
\]
where $F(t)$ is the value at iteration $t$, $t_{max}$ is the total number of iterations, $\alpha_0$ is the initial value we anneal from and $\alpha_1$ is the final value we anneal to.

\subsection{Snapshot Ensembles}

Snapshot Ensembling is a popular low-cost ensemble learning approach.
Snapshot Ensembles work by training only a single model and saving checkpoints of that model's state over time.

Training is broken up into several cycles, where for each cycle the learning rate is decayed from a large value to a small value.
The large rates encourage the model to move further in the parameter space before converging to a local optima which is then saved \cite{huang2017snapshot}.
This process then repeats, with the learning rate resetting to the large value at the beginning of each new cycle.

The total number of ensemble members that are created for the final ensemble is governed by the cycle length and total training budget.
The cycle lengths need to be long enough to allow for single models to optimize well, however this can limit the potential number of ensemble members created.
Within each cycle, the learning rate is decayed with cosine annealing, and the learning rate at any given iteration can be described with:

\[
a(t) = F(mod(t - 1, \lceil T/M \rceil))
\]

\noindent where $a(t)$ is the learning rate at iteration $t$, $F()$ is the cosine annealing function, $T$ is the total number of iterations, and $M$ is the cycle length used for each model.

Snapshot Ensembles are able to create ensembles with good performance without requiring any more computation than a single model would have. However, snapshots taken early in the training process tend to perform worse and snapshots taken later in the training tend to be highly correlated and less diverse, thus resulting in less effective ensembles.

To provide context on the relative performance of these approaches, we include Table 1 which presents results comparing Accuracy (ACC), Negative Log Likelihood (NLL) and Expected Calibration Error (ECE) for both Prune and Tune Ensembles and Snapshot Ensembles as well as several other recent and low cost ensemble methods for CIFAR-10 and CIFAR-100. All methods are trained with roughly the same computational budget and use a standardized WideResNet-28-10 convolutional model.
Prune and Tune Ensembles outperform all other methods with an accuracy of 96.48 on CIFAR-10 and 82.7 on CIFAR-100 when accounting for a fixed training budget. Snapshot Ensembles result in accuracies of 96.27 and 82.1 respectively.
A full table of results, including computational costs can be found in the appendix.

\begin{table}[t]
\caption{Comparison between Low Cost Ensembles using WideResNet-28-10 Architecture. $M$ refers to the total number of models in the final ensemble. Besides Prune and Tune (PAT) and Snapshot Ensembles, other methods include: Dropout \cite{hinton2012improving}, TreeNet \cite{lee2015m}, Batch Ensembles \cite{wen2020batchensemble}, Fast Geometric Ensembles \cite{garipov2018loss}, and Multiple-Input Multiple-Output(MIMO) \cite{havasi2021training}.  * indicates published results taken from \cite{whitaker2022prune ,havasi2021training, liu2021freetickets}.}
\begin{center}
\begin{tabularx}{\columnwidth}{X ccc ccc}
\toprule
 $~$    & \multicolumn{3}{c}{CIFAR-10} & \multicolumn{3}{c}{CIFAR-100} \\
\midrule
Method & ACC & NLL  & ECE  & ACC  & NLL  & ECE \\
\midrule
Dropout*         &   95.9    &  0.15 &  	0.024   & 79.6 & 0.83 & 0.05 \\ 
Treenet (M=3)*   & 95.9      & 0.25  & 0.018     & 80.8 & 0.77 & 0.05 \\
Batch (M=4)*  & 96.2      & 0.14  & 0.02      & 81.5 & 0.74 & 0.05 \\
SnapShot (M=5)       & 96.27      & 0.13  & 0.02      & 82.1 & 0.66 & 0.04 \\
FGE (M=12)       & 96.35      & 0.13  & 0.02      & 82.3 & 0.65 & 0.04 \\
MIMO (M=3)*       & 96.4      & 0.13  & 0.01      & 82.0 & 0.69 & 0.02 \\
PAT (M=6)       & 96.48     & 0.11 & 0.005     & 82.7 & 0.63 & 0.01 \\
\bottomrule
\end{tabularx}
\end{center}
\end{table}

\subsection{Output Diversity}

Diversity has long been known to be an important consideration in ensemble learning \cite{ueda1996generalization, brown2005diversity, kuncheva2003measures}.
This is often explained in ensemble literature with an example of the bias-variance decomposition of the mean squared error (MSE) \cite{ueda1996generalization, brown2005diversity}. Given a model's prediction $f$  and a true target value from an unknown test distribution $y$, the mean squared error is defined to be the expectation of the squared distance between the models predictions and the true target distribution. Bias is the difference between the expectation of the model and the true targets and variance is the squared difference between the models predictions and its mean.
\begin{gather*}
bias = E[f] - y \\
var = E[(f - E[f])^2] \\
MSE = E[(f - y)^2] = (E[f] - y)^2 + E[(f - E[f])^2] \\
MSE = bias^2  + var
\end{gather*}

Given an ensemble of $M$ equally weighted estimators, the decomposition can be further extended to produce the bias-variance-covariance decomposition \cite{brown2005diversity}.
\begin{gather*}
    \overline{bias} = \frac{1}{M} \sum_i(E[f_i] - y) \\ 
    \overline{var} = \frac{1}{M} \sum_i E[(f_i - E[f_i])^2]\\
    \overline{covar} = \frac{1}{M(M-1)} \sum_i \sum_{j \neq i} E[(f_i - E[f_i])(f_j - E[f_j])] \\
    MSE = \overline{bias^2} + \frac{1}{M} \overline{var} + (1 - \frac{1}{M}) \overline{covar}
\end{gather*}

Generalization error for single models relies upon the optimization of both bias and variance, where the tuning of a model towards high bias can cause it to miss important features and the tuning of a model toward high variance can cause it to be highly sensitive to noise. When the decomposition is extended to an ensemble, the generalization performance additionally depends on the covariance between models. Ideally, ensemble methods that prioritize diversity will be able to reduce covariance without increasing the bias or variance of individual models \cite{ueda1996generalization, brown2005diversity}.

In a classification context, there is no standardized analog to the bias-variance decomposition \cite{brown2005diversity}. 
Instead, several metrics have been introduced as a means to quantify diversity in classification ensembles \cite{kuncheva2003measures}. The most popular of which are the Kullback-Leibler Divergence and Prediction Disagreement Ratio \cite{fort2020deep, havasi2021training, liu2021freetickets}. 

Kullback-Leibler Divergence, also known as relative entropy, approximately measures how different one probability distribution is from one another. This operates on the output probabilities of each ensemble member and the average is measured over all pairwise combinations.
\[
d_{KL}(f_1, f_2) = \frac{1}{N} \sum_{i=1}^N f_1(x_i) \log \left( \frac{f_1(x_i)}{f_2(x_i)} \right)
\]
where $N$ is the number of test samples and $f_i(x_i)$ is the output probabilities for a given model $f_i$ and test sample $x_i$.

Prediction Disagreement Ratio (PDR) instead measures the differences between only the predicted class instead of differences between the full output distributions.
\[
d_{PDR}(f_1, f_2) = \frac{1}{N} \sum_{i=1}^N argmax \ (f_1(x_i)) \neq argmax \ (f_2(x_i))
\]
where $N$ is the number of test samples and $argmax(f_i(x_i))$ is the predicted class label for model $f_i$ and test sample $x_i$.

Table II presents results from several low-cost ensemble papers on a benchmark CIFAR-10 experiment with a total training budget of 200 epochs \cite{liu2021freetickets, havasi2021training, whitaker2022prune}. The results indicate that output diversity metrics do not directly correlate with the most accurate ensembles. For example, Dense Ensembles have a prediction disagreement ratio or 0.032 and a KL-divergence of 0.086 with an accuracy of 96.6. These diversity metrics are lower than Prune and Tune (PAT) Ensembles and Dynamic Sparsity Training (DST) Ensembles despite having a higher accuracy. It's clear that diversity metrics alone are not sufficient to fully analyze the generalization capabilities of different methods.

\begin{table}[t]
\caption{Prediction Disagreement Ratio (PDR) and KL divergence between ensemble members on CIFAR-10 with WideResNet-28x10. Results reported from \cite{liu2021freetickets, havasi2021training, whitaker2022prune}}

\begin{tabularx}{\columnwidth}{X c c c}
\toprule
Methods & $d_{PDR}$ $\uparrow$ & $d_{KL}$ $\uparrow$ & Acc $\uparrow$ \\
\midrule
Treenet & 0.010 & 0.010 & 95.9\\
BatchEnsemble & 0.014 & 0.020 & 96.2 \\
EDST Ensemble & 0.026 & 0.057 & 96.4 \\
MIMO & 0.032 & 0.081 & 96.4 \\
Dense Ensemble & 0.032 & 0.086 & 96.6 \\
DST Ensemble & 0.035 & 0.095 & 96.4 \\
Prune and Tune Ensemble & 0.036 & 0.090 & 96.5 \\
\bottomrule

\end{tabularx}
\end{table}

\section{Interpretable Diversity Analysis}

The efficacy of low-cost methods often rely on a trade-off between training cost, model accuracy, and member diversity. However, the current state of diversity analysis focuses heavily on output diversity, which gives little insight into how networks actually develop and represent diversity.

This chapter explores diversity by visualizing the feature spaces of child networks generated in Prune and Tune Ensembles and compares them to checkpoint models in Snapshot Ensembles. These networks offer an excellent foundation for exploring diversity for a number of reasons. 

1) Prune and Tune Ensembles demonstrate excellent empirical results and output diversity, despite the fact that each child network inherits parameters from a shared parent and is tuned for only a small number of epochs. Visualizing the feature representations of these models can provide an explanation as to how and why these methods produce robust ensembles.

2) Child networks in Prune and Tune Ensembles and Snapshot Ensembles are derived from an identical parent. We can look at the index of the same neuron in two different networks or checkpoints and visualize how feature representations diverge and change over time and space.

3) Sparsity has long been an important topic in deep neural network research. Trained networks can have a significant number of parameters removed and continue to maintain, or even exceed the original network's accuracy with little additional training. Visualizing how sparsity affects the underlying knowledge representations is an important area for better understanding the behavior of deep neural networks.


\subsection{Feature Visualization}

\begin{figure*}
\centering
\includegraphics[width=\textwidth]{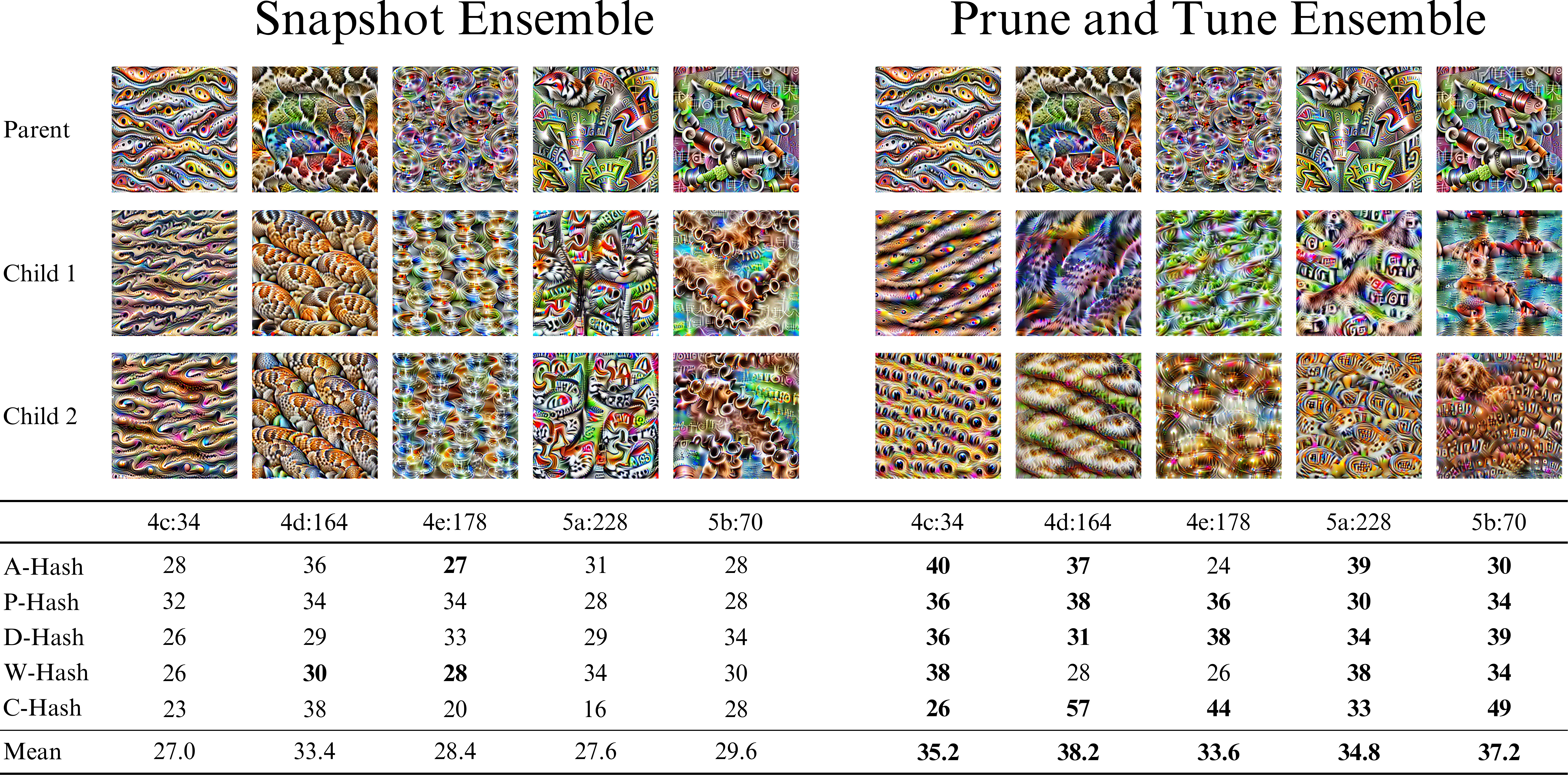}

\caption{Feature visualizations of a random selection of neurons in child networks from a Snapshot Ensemble and a Prune and Tune Ensemble. Each column is labeled with (layer index:channel index) to identify the location within the network of the randomly selected neurons. Values reported are the hash distances between the two child network visualizations. Bold values represent greater distance between representations. The sparsity from Prune and Tune Ensembles results in much more diverse feature representations than two subsequent checkpoints in Snapshot Ensembles.}
\end{figure*}

Feature visualization uses optimization to create images that maximize or minimize activations of specific parts of the network. This is done by first initializing an image with random noise. This image is then fed into a fixed network and gradient descent is used to update the pixel values of the input image. The resulting images display the types of patterns a particular neuron or channel responds strongly to.


Various forms of regularization have been shown to improve image quality, including frequency penalization, transformation robustness, and learned priors (such as color distribution) \cite{olah2017feature}. Additionally, images are preconditioned by decorrelating and whitening the input such that gradient descent happens in Fourier Space, with frequencies scaled to have the same energy \cite{olah2017feature}. The 2D Fourier Transform for an $x$-by-$y$ image $X$ is defined as:
\[
\mathcal{F}(u,v) = \sum_{i=0}^{x} \sum_{j=0}^{y} e^{-2 \pi i / x} e^{-2 \pi i / y} X_{i, j}
\]
%

We then aim to maximize the output activations of a specific convolutional filter by optimizing the following objective:
\[
F(X) = -\sum_{x,y}h_{n,x,y,z}(X)
\]
where $h$ is the output of a neuron, $X$ is the input image, $n$ is the layer, $z$ is the channel, and $x$ and $y$ are the spatial positions of the neuron within a channel. In order to generate an image that minimizes neuron activation, the sign is flipped for the objective. The image is then optimized using a standard gradient descent based optimizer:
\[
X_{n+1} = X_n - \eta \nabla F(X_n)
\]
where $\eta$ is the learning rate and $\nabla$ is the gradient of the objective function with respect to the input image.

\subsection{Saliency Maps}

Saliency Maps visualize the gradient of a prediction with respect to the input image, highlighting the parts of the input that a network responds strongly to. This is done by first performing a forward pass through the network with a given input image. The saliency is computed by setting all outputs of the non-predicted class to 0 and back propagating the predicted class score back through to the inputs \cite{avanti2017saliency}. The resulting gradient is then normalized and plotted using a heatmap, where larger values correspond to more salient parts of the input space.
\[
G(X) = \frac{\partial Y_c}{\partial X}
\]
where the saliency map $G(X)$ is the gradient of the output of the predicted class label $Y_c$ with respect to the input image $X$.

SmoothGrad was later introduced as an extension to the method described above that reduces noise in the visualizations by averaging the gradient over several Gaussian perturbed inputs \cite{smilkov2017smoothgrad}.
\[
SG(X) = \frac{1}{N} \sum_{i=1}^N \frac{\partial Y_c}{\partial X + \gamma_i}
\]
where $SG(X)$ is the SmoothGrad output for an input image $X$ and $Y_c$ is the output of the predicted class label $c$. The input $X$ is perturbed with Gaussian noise sampled from a normal distribution $\gamma_i \sim \mathcal{N}(0, \sigma^2)$.

\subsection{Quantifying Visual Diversity}

\begin{figure*}
\centering
\includegraphics[width=\textwidth]{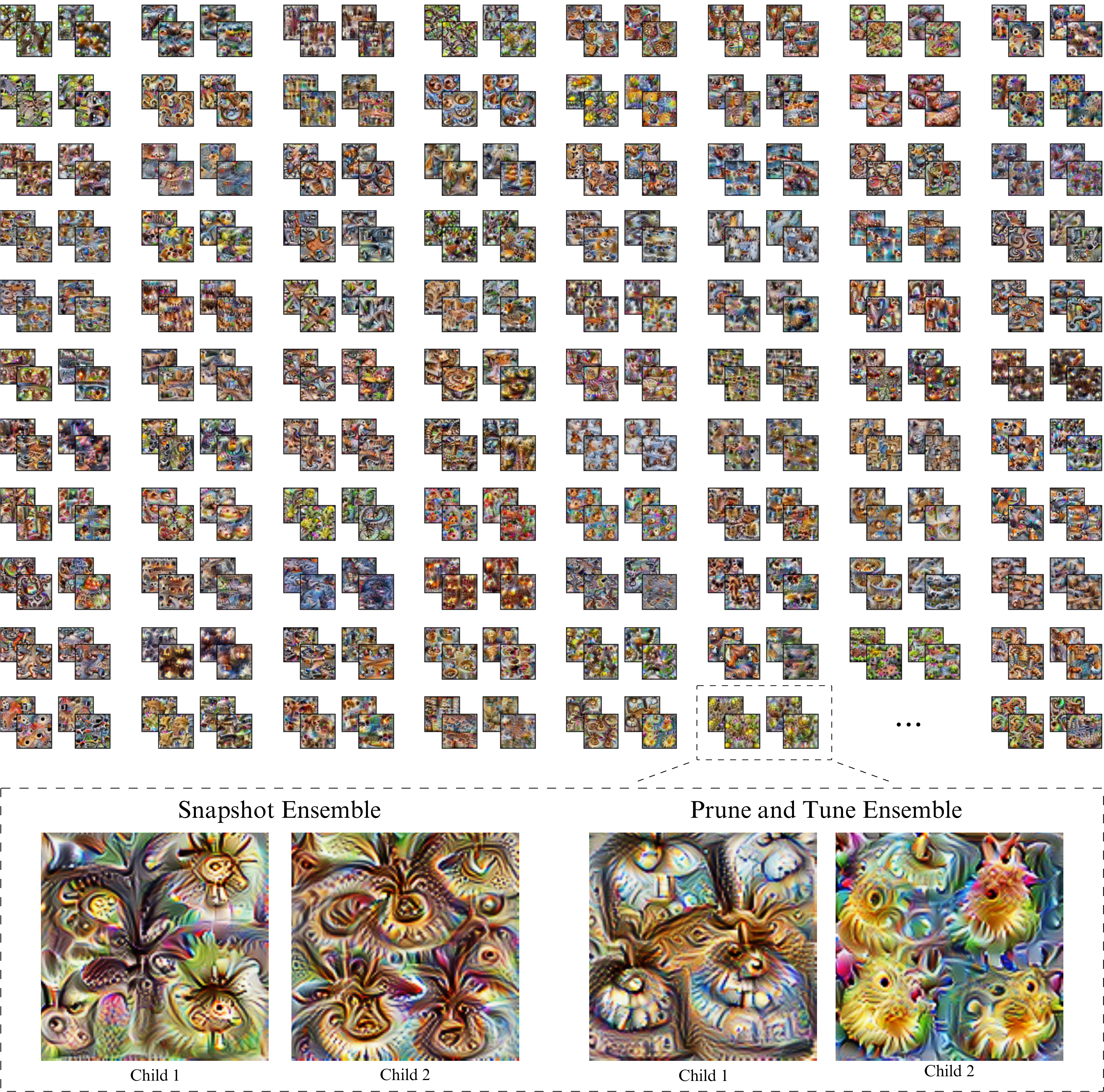}

\vspace{0.2in}

\begin{tabularx}{\textwidth}{X c c c c c}
\toprule
Method & Average Hash & Perceptual Hash & Difference Hash & Wavelet Hash & Color Hash \\
\midrule
Snapshot Ensemble & $30.85 \pm 0.27$ & $30.77 \pm 0.25$ & $31.55 \pm 0.28$ & $30.94 \pm 0.28$ & $27.56 \pm 0.39$ \\
Prune and Tune Ensemble & $\textbf{31.16} \pm 0.28$ & $\textbf{31.19} \pm 0.25$ & $\textbf{31.60} \pm 0.27$ & $\textbf{31.17} \pm 0.27$ & $\textbf{32.74} \pm 0.51$ \\
\bottomrule
\end{tabularx}

\vspace{0.1in}

\caption{We measure the hash difference between sibling network feature representations using a variety of perceptual hash techniques. Results are the mean distances between all 1024 neurons in the final convolutional layer before the linear classifier. Each cluster of 4 filter visualizations are from the same neuron in different child networks that are each derived from a shared parent network. The left overlapping filters are from two successive snapshot checkpoints. The right overlapping filters are from two prune and tune children. \textbf{Bolded} values correspond to greater distance and greater diversity between child filter representations.}
\end{figure*}

The most common image similarity metrics are generally used to measure image quality as a response to noise, corruption, or compression against a ground truth image. These include the mean square error, peak signal-to-noise ratio, structural similarity index, spatial correlation coefficient, spectral angle mapper, and universal image quality index. These metrics tend to be highly sensitive to minute differences and far too noisy when applied to images that contain different content.

We instead introduce a robust image hash similarity based approach to quantifying diversity. Image hashing algorithms compress images into binary bit strings such that images that are visually similar will result in hashes that are similar. The bits of the image hash are spatially significant, so the similarity between two hashes can be described by the Hamming Distance, which is the number of positions in which the bits of two hashes differ. Images that are more similar will therefore have a lower Hamming Distance than images that are diverse \cite{zauner2010implementation}.


We use several popular image hashing algorithms for measuring the distances between feature visualizations, including: average hash, perceptual hash, difference hash, wavelet hash, and color hash \cite{buchner2021imagehash}. Average hash, difference hash, wavelet hash and perceptual hash all start by resizing a given image to an 8x8 pixel square and converting it to grayscale. The hash is then constructed by assigning a 1 or 0 for each pixel in the block according to some heuristic. Average Hash assigns a 1 if the pixel value is greater than the mean pixel value. Perceptual Hash does the same except all values are converted to the frequency domain using a discrete cosine transformation first. Wavelet hash uses a discrete wavelet transform instead of a discrete cosine transform. Difference hash can be described as a gradient hash where the difference between pixels values is compared to the mean difference between pixel values. Finally, color hash maintains color information by skipping the grayscale step and instead using the hue, saturation, and value space of an image separately.

\section{Experiments}

We conduct a large scale comparison between child networks in a Snapshot Ensemble and a Prune and Tune Ensemble. We start with an open source Inception network pre-trained on ImageNet \cite{torchvision}. Following the hyperparameters given in the Snapshot Ensemble paper, we continue training of the pre-trained network for two cycles of 40 epochs in length with a cyclic cosine-annealing decay schedule with an initial learning rate of 0.1 and a final learning rate of 1e-5 \cite{huang2017snapshot}.

We use the same pre-trained network as the parent for the Prune and Tune Ensemble as well. Two child networks are created using anti-random pruning with an unstructured sparsity target of 50\%. Each child is fine tuned for 40 epochs using a one-cycle learning rate schedule with a max learning rate of 0.1 and a final learning rate of 1e-5. We choose these hyperparameters to match the computational budget defined in Snapshot Ensembles \cite{huang2017snapshot}.

\begin{figure}
    \includegraphics[width=\columnwidth]{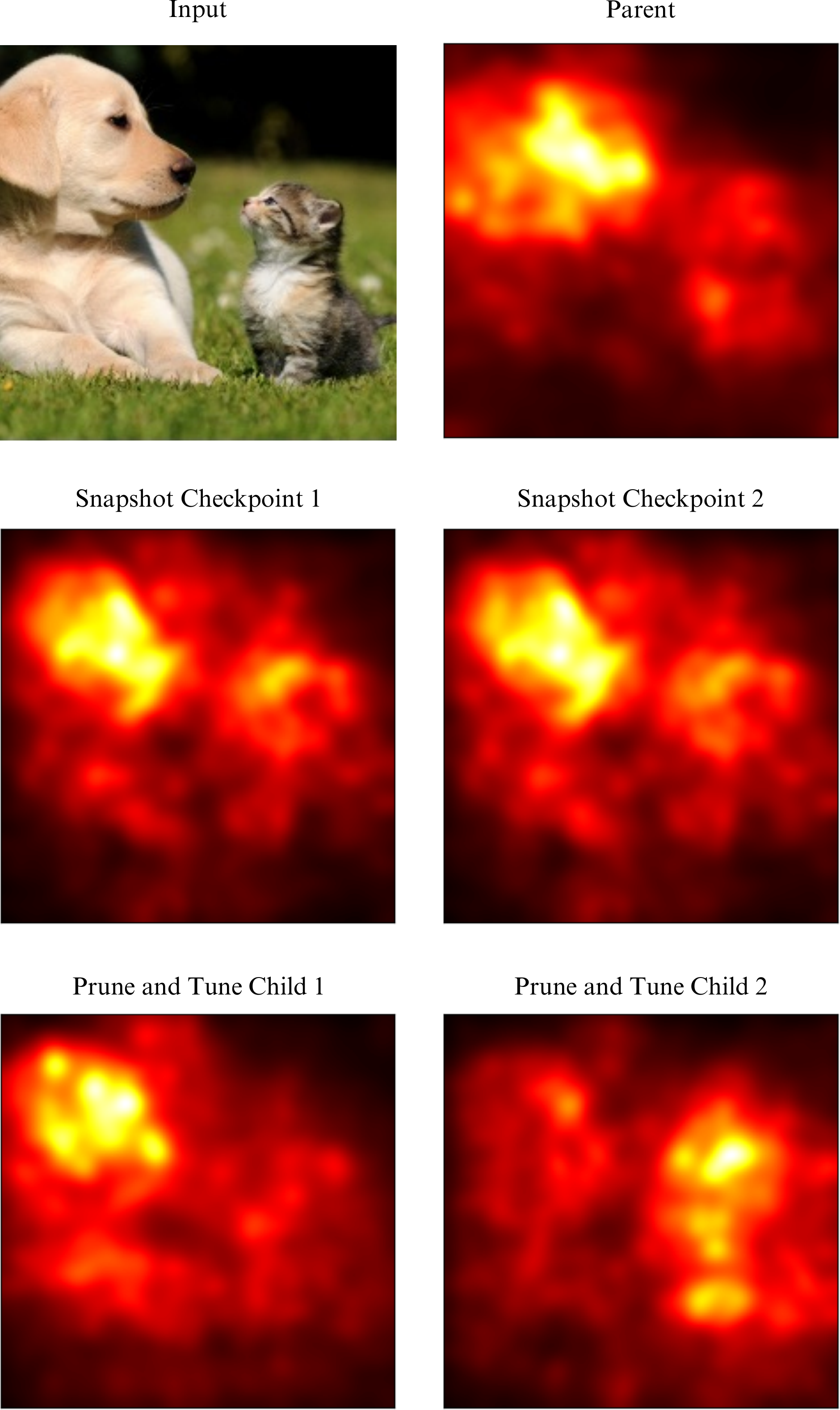}
    
    \vspace{0.2in}
    
    \small
    \begin{tabularx}{\columnwidth}{X @{ }c @{ }c @{ }c @{ }c @{ }c}
    \toprule
     & RMSE & A-Hash & P-Hash & D-Hash & W-Hash \\
    \midrule
    SSE & $0.06^{\pm 0.001}$ & $10.01^{\pm 0.04}$ & $18.12^{\pm 0.04}$ & $10.00^{\pm 0.04}$ & $15.13^{\pm 0.04}$ \\
    PAT & $\textbf{0.07}^{\pm 0.001}$ & $\textbf{11.13}^{\pm 0.04}$ & $\textbf{19.95}^{\pm 0.04}$ & $\textbf{11.03}^{\pm 0.04}$ & $\textbf{15.97}^{\pm 0.04}$ \\
    \bottomrule
    \end{tabularx}

    \vspace{0.1in}

    \caption{Mean distances between Snapshot Ensemble (SSE) and Prune and Tune Ensemble (PAT) saliency maps for all 50,000 samples in the ILSVRC2012 ImageNet validation set. Also included is an example of how saliency maps can be useful with ambiguous input. Included are SmoothGrad saliency maps for a shared parent network, two snapshot ensemble checkpoints, and two prune and tune child networks. Both snapshot checkpoints focus on the dog while the prune and tune children split their attention between the dog and the cat. The diversity encouraged by prune and tune ensembles helps to reduce bias present in the parent network.}
\end{figure}

Using the child networks constructed above, we create feature visualizations of identical neurons in each of the networks \cite{kiat2021lucent, tensorflow2021lucid}. Feature visualizations are optimized using ADAM with a learning rate of 0.05 for 1024 steps. Several random data augmentations are applied at each step, including: jittering by up to 8 pixels, scaling by a factor between 0.95 and 1.05, rotating by an angle between -5 and 5 degrees, and jittering a second time by up to 4 pixels.

We then compute 64 bit image hashes for each visualization (average hash, perceptual hash, difference hash, wavelet hash, and color hash) and report the Hamming Distance between each pair of child network visualization hashes.

We first present feature visualizations from several neurons selected at random from different layers within each network. Figure 2 displays these sample visualizations where each column corresponds to the same neuron in each of the different networks. Prune and Tune children often focus on different elements present in the parent network's visualization, while Snapshot checkpoints appear to be much more highly correlated with the parent. The table in Figure 2 supports this, with the Hamming distance between each pair of image hashes being more distant than those created with Snapshot Ensemble checkpoints.

We extend these feature visualizations by analyzing every channel in the final convolutional layer. Thus, for each pair of child networks, 1024 feature visualizations are created for each. All of these visualizations are hashed using the algorithms described above and then the distance between sibling network hashes are measured. These final neurons are especially relevant for a comprehensive comparison as they are the last stop before being fed into the fully-connected classifier layers. They most closely illustrate how each child network represents concepts and the differences over all the neurons in this layer will provide a comprehensive overview on the diversity between network representations. 

Figure 3 displays a subset of these visualizations along with a table that reports the mean Hamming distances between each pair of child networks over all feature visualization hashes. For every hashing algorithm we see more diversity in the Prune and Tune representations than in the Snapshot Ensemble.

We then explore how each ensemble method interprets input by creating saliency maps for each of the 50,000 samples in the ImageNet ILSVRC2012 validation set \cite{deng2009imagenet}. Figure 4 reports the average Root Mean Square Error, Average Hash, Perceptual Hash, Difference Hash, and Wavelet Hash distance over all 50,000 images between each pair of child networks.

Figure 4 also includes a visualization of how saliency maps can be useful in analyzing ensemble diversity when dealing with ambiguous inputs. We feed an image containing a puppy and a kitten to each child network in the Snapshot and Prune and Tune Ensemble and compute their saliency maps using SmoothGrad. The parent network and the Snapshot checkpoints focus primarily on the puppy in the input image. The Prune and Tune children split their attention, where one child focuses on the puppy and the other on the kitten. This diversity can be valuable in building robust ensembles as techniques that encourage diverse representations can help to reduce bias (focus on the puppy) that is inherent in the parent network.

\section{Conclusions}

Diversity is an important property of robust neural network ensembles. However, traditional measures of diversity focus only on model outputs and give little insight into how ensemble members represent knowledge differently within the model. We introduce a combination of interpretability methods and perceptual hashing as an effective approach for qualitatively analyzing diversity and measuring the similarity between representations.

Diversity is especially important for low-cost ensemble methods as they tend to share information between members in order to reduce computational cost. In particular, temporal ensembles and evolutionary ensembles create members that inherit parameters and network structure from previous iterations, which allows for meaningful comparisons between the same neurons of different networks.

Our experiments explore how feature visualization and saliency maps could be useful in comparing the diversity between members in two low-cost ensemble methods, Snapshot Ensembles and Prune and Tune Ensembles. We find that the sparsity and unique network topology of Prune and Tune children is highly effective for encouraging diverse feature representations when compared to Snapshot Ensembles. Anti-random pruning ensures that the convolutional filters are geometrically opposed which results in convergence to unique and diverse optima. Despite the significant pruning that these children undergo, Prune and Tune Ensembles maintain high classification accuracy \cite{whitaker2022prune}.

Interpretability methods are making deep neural networks more accessible and understandable, and we believe the introduction of these methods to ensemble learning can provide better insights into diversity and aid in the construction of more robust low-cost ensembles.

\bibliographystyle{plain}
\bibliography{bibliography}

\end{document}